\documentclass{bmvc2k}
\usepackage{hyperref}

\title{Difflare: Removing Image Lens Flare with Latent Diffusion Models}

\addauthor{Tianwen Zhou}{tianwenzhou0521@ieee.org}{1}
\addauthor{Qihao Duan}{7hao.duan@gmail.com}{1}
\addauthor{Zitong Yu*}{yuzitong@gbu.edu.cn}{1}

\addinstitution{
School of Computing and Information Technology, Great Bay University\\
}

\runninghead{Zhou et al.}{Removing Image Lens Flare with
Latent Diffusion Model}

\def\etal{\emph{et al}\bmvaOneDot}

\begin{document}

\maketitle

\begin{abstract}
The recovery of high-quality images from images corrupted by lens flare presents a significant challenge in low-level vision. Contemporary deep learning methods frequently entail training a lens flare removing model from scratch. However, these methods, despite their noticeable success, fail to utilize the generative prior learned by pre-trained models, resulting in unsatisfactory performance in lens flare removal. Furthermore, there are only few works considering the physical priors relevant to flare removal. To address these issues, we introduce Difflare, a novel approach designed for lens flare removal. To leverage the generative prior learned by Pre-Trained Diffusion Models (PTDM), we introduce a trainable Structural Guidance Injection Module (SGIM) aimed at guiding the restoration process with PTDM. Towards more efficient training, we employ Difflare in the latent space. To address information loss resulting from latent compression and the stochastic sampling process of PTDM, we introduce an Adaptive Feature Fusion Module (AFFM), which incorporates the Luminance Gradient Prior (LGP) of lens flare to dynamically regulate feature extraction. Extensive experiments demonstrate that our proposed Difflare achieves state-of-the-art performance in real-world lens flare removal, restoring images corrupted by flare with improved fidelity and perceptual quality. The codes will be released soon.

\end{abstract}

\section{Introduction}
\label{sec:intro}
\begin{figure}
\label{fig1}
\begin{tabular}{cc}
\bmvaHangBox{\includegraphics[width=3cm]{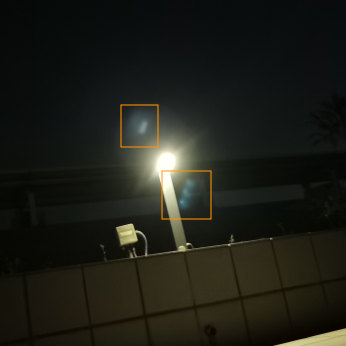}\\}
\bmvaHangBox{{\includegraphics[width=3cm]{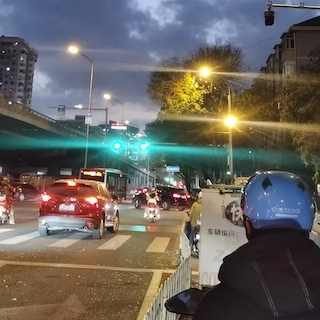}}}&
\bmvaHangBox{\includegraphics[width=6.5cm]{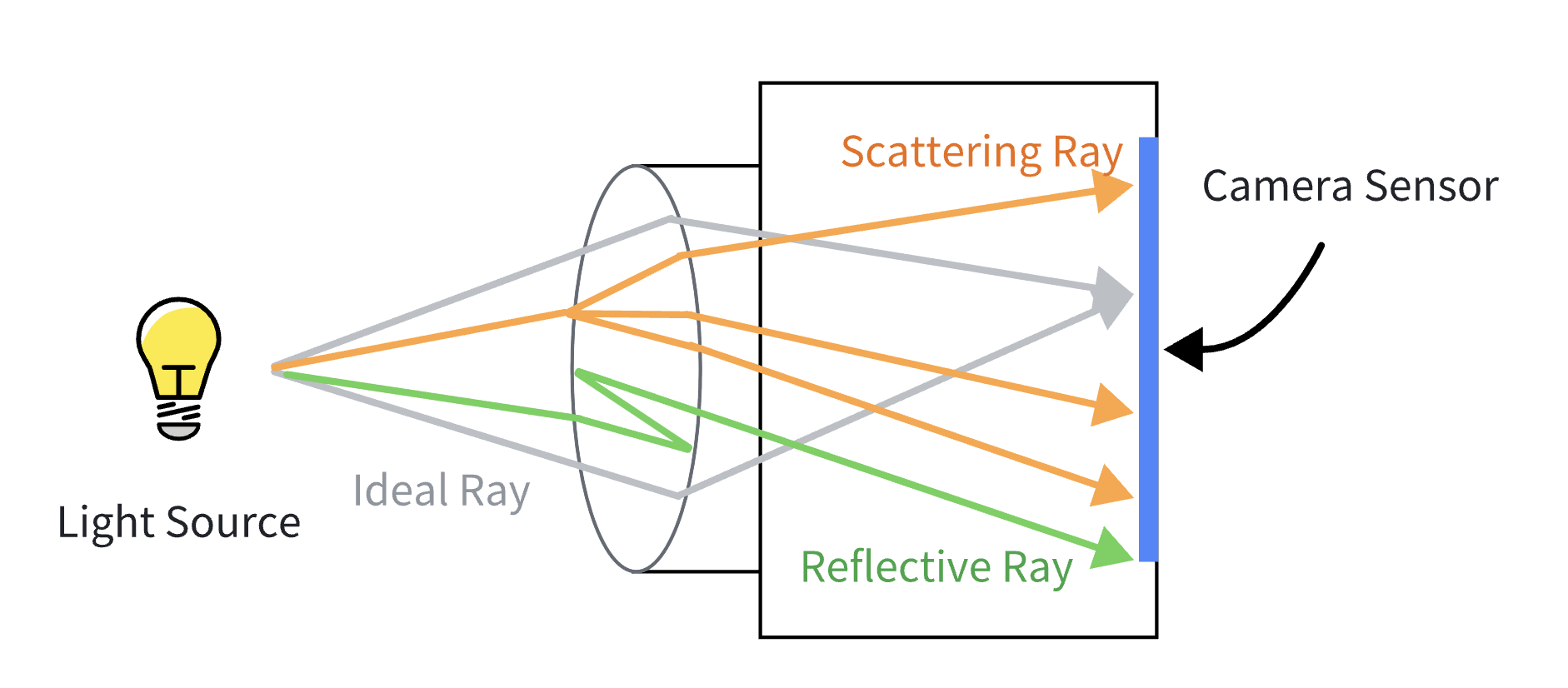}}\\
(a)&(b)
\end{tabular}
\caption{\small{(a) Real-world captured RF and SF; (b) Diagram of the origin of lens flare. Ideally, rays from the point light source incident to the lens are intended to be focused at a single point on the camera sensor (gray rays). However, due to the dust, wear and scratches on the surfaces of the lenses and the reflection between the air-glass interface of the lens, the incident rays might be scattered or reflected to unexpected locations, leading to unwilling artifacts on the captured image. }}
\label{fig:flare}
\vspace{-0.8em}
\end{figure}

Lens flare is a form of local degradation present in images captured by various cameras, which significantly diminishes image quality and affects real-world applications, such as autonomous driving \cite{ZHANG2022e11570}.
The two primary types of lens flare are Reflective Flare (RF) and Scattering Flare (SF). RF is often caused by multiple reflections at the air-glass interface of the lens \cite{10.1145/2010324.1965003}, which often manifest as polygons and circles on the captured image. SF occurs when light scatters on the surface of lenses due to scratches, fingerprints, or dust, resulting in radial line patterns, as illustrated in \hyperref[fig1]{Figure 1(a)}. Additionally, this phenomenon is more pronounced at night due to the presence of multiple artificial lights \cite{dai2022flare7k}.

Current approaches \cite{dai2022flare7k}\cite{dai2023flare7kpp}\cite{qu2024harmonizing} to lens flare removal involve training learning-based methods from scratch . They commonly use semi-synthetic flare-corrupted images as an input guidance to restore the corresponding flare-free images. While these methods have made noticeable progress, they have failed to leverage the optical and generative priors captured in pre-trained generative models (PTGM). Moreover, they demand substantial computational resources for training from scratch, resulting in suboptimal outcomes and reduced robustness in real-world scenarios. Such drawbacks have led to sub-optimal results and less robustness in real-world scenarios.

While there have been no previous attempts to address these issues in lens flare removal, several approaches have been proposed to use PTGM as generative priors for other global image restoration (IR) tasks. Wu \etal \cite{wu2023ridcp}  pre-trained a VQ-GAN \cite{esser2020taming} on high quality haze-free images to leverage the high-quality codebook prior in image dehazing task. Wang \etal \cite{wang2023exploiting} employed a time-aware encoder to fine-tune pre-trained diffusion models (PTDM) for injecting guidance into image super-resolution tasks. However, these methods can only achieve satisfactory results in restoring global degradation, such as haze and low-resolution, but often fail to maintain fidelity in non-corrupted areas when addressing local degradation, such as lens flare.

In this work, we present a novel paradigm for image lens flare removal named Difflare, motivated by the challenges outlined previously. Specifically, we harness the robust generative priors of natural images captured in PTDM by maintaining the original parameters and incorporating structural conditions to guide lens flare removal via our proposed Structural Guidance Injection Module (SGIM). Unlike previous approaches that start training from scratch, we simply fine-tune PTDM and conduct training in latent space, substantially reducing the computational cost compared to training from scratch. To address information loss resulting from latent compression and to maintain fidelity in flare-free areas, we introduce the Adaptive Feature Fusion Module (AFFM), which fuses the encoded feature with the decoded feature as a residual with a modified attention mechanism under the guidance of the Luminance Gradient Prior (LGP).

To the best of our knowledge, our method is the first to perform lens flare removal in latent space with a focus on local degradation removal. Extensive experiments demonstrate that our method surpasses the current state-of-the-art (SOTA) methods in both fidelity and perceptual quality.

\vspace{-0.8em}
\section{Related Works}
\label{sec:relatedworks}
\subsection{Lens Flare Removal}

The objective of lens flare removal methods is to restore a clear image from one that has been corrupted by flare. Numerous methods have been developed for the removal of lens flare from images. A common approach involves masking lenses with Anti-Reflective (AR) coatings to reduce reflections between them. However, this method lacks support for post-capture processing and requires meticulous design. Therefore, employing automated algorithms to remove lens flare from corrupted images is preferable. Post-capture processing methods \cite{Chabert2015AutomatedLF} predominantly rely on the optical traits \cite{5559015} of lens flare to enable automated detection and removal. However, such methods may suffer from poor robustness across various scenarios and types of corruption.

{\bf Learning-based Methods.} The rapid progress of deep learning algorithms has significantly contributed to the success of image restoration (IR) using deep learning techniques. Several endeavors have been made to apply learning-based approaches to lens flare removal. Dai \etal \cite{dai2022flare7k} trained a lens flare removal network from scratch, using UNet \cite{ronneberger2015unet}, Restormer \cite{Zamir2021Restormer}, Uformer \cite{wang2021uformer} and MPRNet as baseline architecture respectively, and found that Uformer can achieve the best performance. Zhang \etal \cite{10208804} proposed swin-transformer in Fourier space as the network backbone. However, these methods tend to simultaneously remove both the light source and the lens flare from the flare-corrupted image, which deviates from the objective of lens flare removal. Hence, Wu \etal \cite{2020How}, Zhou \etal \cite{zhou2023improving} proposed to recover the light source by post-processing the network output with light source mask based on the threshold of brightness of the flare a to add back the light source.

{\bf Lens Flare Dataset.} Given that these methods heavily rely on paired training sets, they necessitate laborious efforts. However, lens flare removal constitutes an inherently ill-posed inverse problem, making it nearly impossible to acquire a sufficient real-world paired training set. Hence, several pioneer works have been proposed to synthesize paired flare-corrupted and flare-free images. Wu \etal \cite{2020How} were the first to propose a purely synthetic paired dataset for flare removal. After that, Dai \etal \cite {dai2022flare7k}  introduced a semi-synthetic benchmark paired dataset named Flare7K, comprising 5,000 SF and 2,000 RF images. This dataset can be readily integrated into any existing natural image dataset to construct a paired training set for lens flare removal. Qiao \etal \cite{9710915} proposed a training pipeline utilizing unpaired flare-corrupted and flare-free images, employing cycle consistency loss \cite{zhu2020unpaired}. In order to synthesize more authentic lens flare, Dai \etal \cite{dai2023flare7kpp} utilized optical center symmetry prior to creating a training set named Flare7K++, effectively distinguishing between the light source and lens flare, thus mitigating the preservation issue of the light source in lens flare removal. Qu \etal \cite{qu2024harmonizing} proposed a data synthesis pipeline guided by the principles of illumination and depth information, driven by the shortcomings of current synthesis methods in producing datasets with a diverse range of background scenes. In our study, we utilize training on the Flare7K dataset \cite {dai2022flare7k} to ensure equitable comparison with existing methodologies. Additionally, we adopt the standard data synthesis pipeline, detailed in \hyperref[sec:experimentalsettings]{Section 4.1}

\subsection{Priors for Image Restoration.}
\label{sec:priors}
Given that image restoration (IR) tasks are perceived as ill-posed inverse problems, it is essential to incorporate natural image priors to constrain the solution space of restored images. The Total Variation (TV) prior \cite{inproceedings} is frequently employed in tasks such as image denoising or deblurring. The Dark Channel Prior (DCP) \cite{5567108} is utilized for real-world image dehazing. Nevertheless, not all image priors can be explicitly formulated analytically. Consequently, numerous methods have been devised to exploit the implicit priors embedded within generative models. Pan \etal \cite{pan2020exploiting} utilize the generative prior within pre-trained BigGAN models to enhance IR tasks. Wu \etal \cite{wu2023ridcp} employed High Quality Priors (HQP) extracted from haze-free images via pre-training a VQ-GAN \cite{esser2020taming} codebook on high-quality haze-free images, aiming to enhance the efficacy of image dehazing. Wang \etal \cite{wang2023exploiting} utilized the generative prior encapsulated in PTDM by incorporating guidance from a low-resolution input to direct PTDM in reconstructing its high-resolution counterpart. Due to the excellent generation performance of Latent Diffusion Models (LDM) \cite{rombach2022highresolution}, priors captured in latent diffusion models are proved to be the most effective among all the generative priors.

In contrast to the aforementioned methods, our approach concentrates on the task of lens flare removal, which is characterized as the removal of local degradation. This particular task necessitates the precise reconstruction of the flare-corrupted area while preserving fidelity in flare-free regions, an aspect that has not received adequate attention in prior research.

\vspace{-0.8em}
\section{Methodology}
\subsection{Overview}
\begin{figure}[t]
\vspace{-0.3em}
\centering
\bmvaHangBox{\includegraphics[width=12cm]{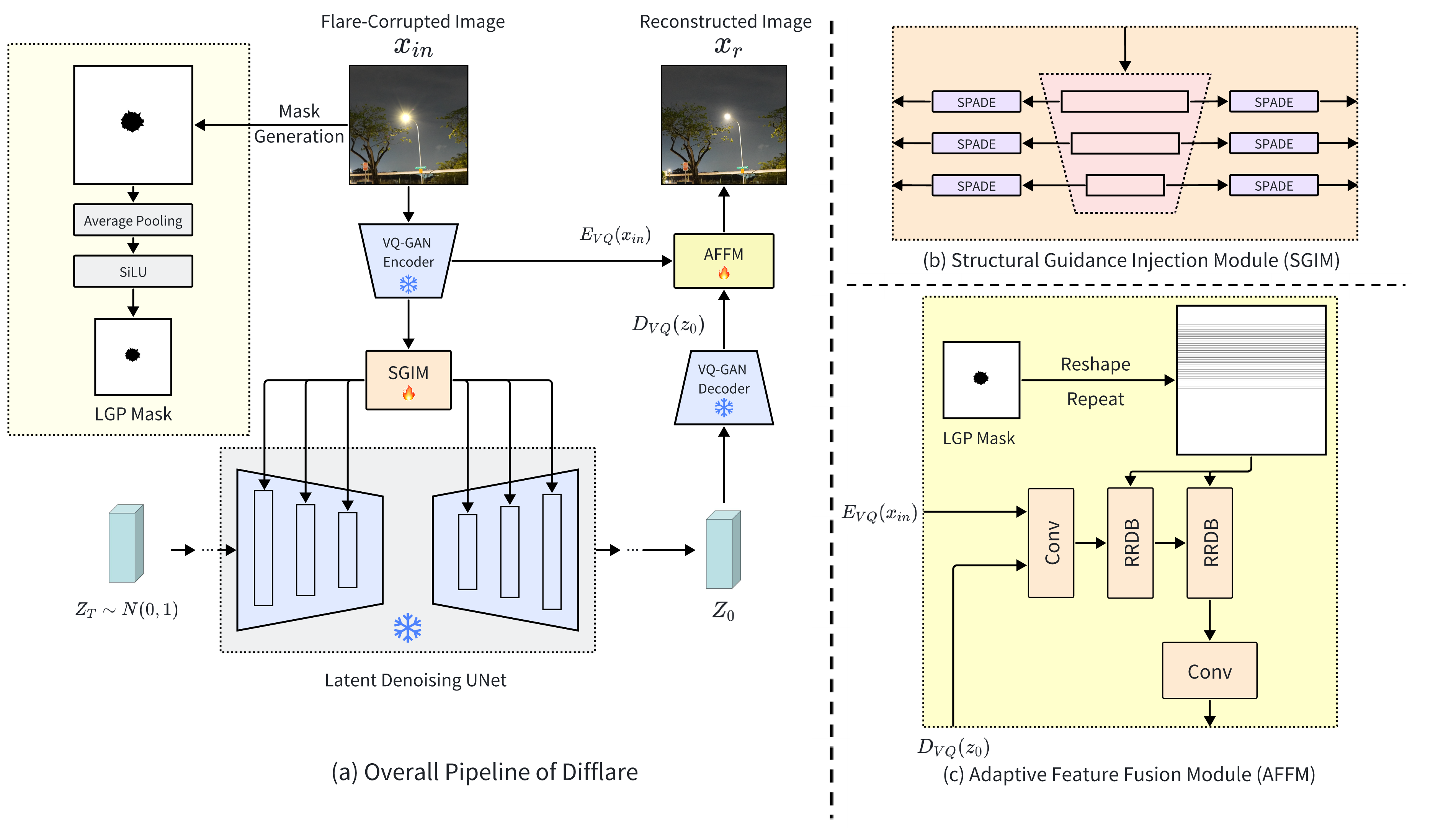}}
\vspace{-0.8em}
\caption{\small{(a) Overview of our proposed Difflare. 
(b) We first utilize the Structural Guidance Injection Module (SGIM) to finetune the frozen Pre-Trained Diffusion Model (PTDM). The multi-scale features extracted by the SGIM is transformed to the corresponding resolution layer of PTDM through Spatially-Adaptive Normalization (SPADE) \cite{park2019SPADE} layers. (c) Additionally, motivated by StableSR \cite{wang2023exploiting}, we introduce a Adaptive Feature Fusion Module (AFFM) to maintain the fidelity between input image and restored image. The AFFM accepts the feature from VQ-GAN encoder and VQ-GAN decoder, and outputs a fusion of both features. The whole process is guided by the Luminance Gradient Prior (LGP) mask via a modification of the self-attention map.}}
\label{fig:overview}
\vspace{-0.8em}
\end{figure}

Our approach, named \emph{Difflare}, seeks to utilize the generative prior inherent in pre-trained latent diffusion models. To reduce the training and fine-tuning expenses of our model, we initially compress the input image using a frozen VQ-GAN \cite{esser2020taming} encoder, enabling flare removal in latent space. In Section 3.2, we introduce our method of fine-tuning PTDM. In \hyperref[sec:AFFM]{Section 3.3}, we elucidate how we maintain fidelity while restoring flare-free images in latent space, guided by the physical priors of lens flare. \hyperref[fig:overview]{Fig.2} illustrates the overview of our work.

\vspace{-0.8em}
\subsection{Injecting Structural Guidance}
\label{sec:SGIM}
To adapt PTDM, a generative text-to-image model, for our lens flare removal task, we must fine-tune it to accept flare-corrupted $x_{in}$ images as conditional inputs. Early finetuning approaches include ControlNet \cite{zhang2023adding}, Lora \cite{hu2021lora} and T2IAdapter \cite{mou2023t2i}, they all follow the same pipeline of freezing the parameters of PTDM while updating a trainable side network. However, these approaches only accept sketches or edges as supplementary guidance. As for our task, we need to generate the reconstructed flare-free image $x_r$ under the guidance of its flare corrupted input counterpart $x_{in}$, which contains far more detail and structural information than sketches or edges. Consequently, we adopt the fine-tuning strategy proposed by Wang \etal \cite{wang2023exploiting}. This involves training a lightweight autoencoder to extract multi-scale information from the input image, guiding the generation process of PTDM. We refer to this module as the Structural Guidance Injection Module (SGIM).

To be specific, we train an autoencoder $E_{SGIM}$ to extract the multi-scale information from the input $z_{in}=E_{VQ}(x_{in})$, representing the corresponding latent vector of $x_{in}$ after encoded by VQ-GAN encoder. $L$ indicates the number of layers of our autoencoder.
This multi-scale feature extraction process effectively captures structural information in a coarse-to-fine manner, producing a set of semantic maps at different resolutions. After the extraction in latent space, we insert the semantic map $\{Fea_i\}_{i=1}^L$ into the residual block of PTDM via multiple Spatially-Adaptive Normalization (SPADE) \cite{park2019SPADE} of different resolutions, which consists of learnable affine transformation layers with semantic map as input. While training the learnable affine transformation layers, we keep other parameters of the PTDM frozen. \\
\begin{equation}
Fea_i = E_{SGIM}(z_{in})_i. \quad Fea_i' = \beta_i\oplus(\gamma_i+1)\otimes Fea_i,  i\in 1,...,L
\end{equation}
in which $\gamma_i,\beta_i$ stands for two trainable convolution layers, $\oplus$ stands for concatenation, and $\otimes$ stands for multiplication.\\
Throughout the training of SGIM, we maintain the parameters of PTDM and VQ-GAN frozen, focusing solely on training the convolution layers above. Such a design can effectively inject multi-scale structural guidance to the generation process of PTDM, while preserving its generative prior, and minimize the training cost of finetuning as well.

\subsection{Adaptive Feature Fusion}
\label{sec:AFFM}


\subsubsection{Luminance Gradient Prior}
\label{sec:LGP}
Explicit and implicit priors hold significance in IR tasks, as mentioned in \hyperref[sec:priors]{Section2.2}. In the lens flare removal task, several explicit priors based on empirical observation or optical knowledge exist. Koreban \etal \cite{5559015} proposed the optical axis symmetric prior according to the optical character of lens flare. Qu \etal \cite{qu2024harmonizing} proposed a depth-related prior to improve the data synthesizing process. In our work, we propose a prior based on the observation of luminance of the flare-corrupted area in YCbCr color space. Specifically, we examine the relationship between the luminance values of an image of size $h\times w$ in YCbCr space. We find that lens flare is corresponding to the areas with significantly higher luminance value, and it is surrounded by a boundary with high luminance gradient. Hence, we generate a Luminance Mask (LM) of size $m = h\times w$ based on this prior and utilize it to guide the training and inference processes of our feature fusion module. Specifically, we choose a threshold value $s$, and we express the value of $i$-th element of the LM as:
\begin{equation}
\label{eq1}
LM_i=
\left \{
\begin{aligned}
1, LM_i &< s \\
0, LM_i &\geq s
\end{aligned}
\right.
  , i=(1,2,...,m)
\end{equation}

\subsubsection{Prior Guided Feature Fusion}
\label{sec:FF}
With the SGIM, our model can effectively produce flare-free images. However, due to the compression of the input flare-corrupted image into a VQ-GAN latent space, inevitable information loss occurs. Also, due to the stochastic sampling process of PTDM, it may also deviate the restored image from the original input. Moreover, unlike other global IR tasks, lens flare removal prioritizes fidelity between input and output images in flare-free areas. Hence, as shown in \hyperref[fig:overview]{Figure 2(c)}, we fuse the features generated by the VQ-GAN encoder $E_{VQ}(x_{in})$ and VQ-GAN decoder $D_{VQ}(z_0)$ through a learnable module composed of several convolution layers and Residual in Residual Dense Block (RRDB) layers, which is proposed by Wang \etal \cite{wang2018esrgan}, the resulting fused feature $F_
{fuse}$ is represented as
\begin{equation}
F_{fuse} = RRDB^n(Conv^m(E_{VQ}(x_{in})\oplus D_{VQ}(z_0)))
\end{equation}
where $\oplus$ represents channel-wise concatenation, and $m,n$ indicates the number of each type of layers.  Furthermore, the aim of the proposed AFFM is to maintain fidelity in flare-free areas. Hence, we attempt to guide the AFFM to focus more on flare-free areas with the Luminance Mask generated with Luminance Gradient Prior (LGP). Specifically, we carefully adjust the self-attention modules in the feature fusion modules, enhancing their awareness of flare-free areas. We first resize the LM calculated by \hyperref[eq1]{Eq.2} to a size of $(h_l, w_l)$ by average pooling and SiLU activation, in which $h_l, w_l$ is equal to the size of the latent vectors in VQ-GAN latent space. Then, we flatten the LM to size $(1, h_l \times w_l)$, and stack $h_l \times w_l$ of LM to form a $(1, h_l \times w_l, h_l\times w_l)$ attention mask LM'. At last, we employ LM' to guide the self-attention calculation by multiplying the weight matrix with LM'. The $attention_m$ of the $m$-th self-attention layer is calculated as follow:
\begin{equation}
Q_m = q_m\cdot W^q, K_m = k_m\cdot W^k, V_m = v_m\cdot W^v,
\end{equation}
\begin{equation}
\textit{attention}_m(Q_m, K_m, V_m)= \textbf{Softmax}[LM'_m\cdot(Q_m\cdot K_m^T)\cdot V_m],
\end{equation}
where $Q_m, K_m, V_m$ represent the Query, Key and Value of self-attention calculation, $q_m, k_m, v_m$ represent the input feature of each self-attention layer, $W^q, W^k, W^v$ indicates the projection matrices. Then, the proposed AFFM is trained with modified self-attention layer under the guidance of the attention mask LM'.
\vspace{-0.8em}
\subsection{Inference Strategy}
During sampling, we initially encode the input image into latent space using our trained AFFM. Then, we sample from random Gaussian noise for 200 DDPM \cite{ho2020denoising} steps in the latent space with guidance from the SGIM and null text-prompt guidance. Moreover, leveraging PTDM priors enables us to capitalize on the innate capabilities of PTDM during the sampling process. Despite training our SGIM with null text prompts, better perceptual quality is attainable when sampling with positive text prompts. Specifically, we facilitate sampling with classifier-free guidance \cite{ho2022classifierfree} of various guidance scale $s$. The overall noise estimation at each time step is
\begin{equation}
\hat{\epsilon}_{\theta}(x_t,c) = (1+s)\cdot\epsilon_{\theta}(x_t,c)-s\cdot\epsilon_{\theta}(x_t,null)
\end{equation}
To enhance the performance of lens flare removal, we utilize the following sets of text prompts to guide the sampling process: \textit{flare free, glare free, (best quality:2), (haze free:2), (very clear:2)}.
\section{Experiments}
\subsection{Experimental Settings}
\label{sec:experimentalsettings}
We select the base version of Stable-Diffusion v2.1\footnote{https://huggingface.co/stabilityai/stable-diffusion-2-1-
base} as the PTDM of our proposed Difflare. We employ training on benchmark lens flare dataset Flare7K \cite{dai2022flare7k}, for fair comparison, we do not include additional training set in \cite{dai2023flare7kpp}. During training, we generate the paired training set on-the-fly, adhering to the synthesizing pipeline described in \cite{dai2022flare7k}. Initially, we randomly sample a background image $B$ from the Flickr24K \cite{zhang2018single} dataset. Subsequently, we randomly select a reflective flare $F_r$, a compound scattering flare $F_s$, and its corresponding light source $L$ from the Flare7K \cite{dai2022flare7k} dataset. We form the flare-free groundtruth $GT$ image as $B\oplus L$, and the flare-corrupted input $x_{in}$ as $B\oplus L\oplus F_r\oplus F_s$, in which $\oplus$ indicates element-wise combination. After that, we adopt the same data augmentation approach as \cite{dai2022flare7k} and randomly crop $GT$ and $x_{in}$ to size $(512,512,3)$. 
Our proposed Structural Guidance Injection Module (SGIM) and Adaptive Feature Fusion Module (AFFM) are trained separately. During the training of SGIM, our model was trained on a 4×RTX4090 GPU with an overall batch size of 192 for 85 epochs. Additionally, we set the text prompt as null during the fine-tuning process. As for training the AFFM, we trained for 11 epochs with an overall batch size of 48.
To evaluate the effectiveness of our proposed method, we assess Difflare on the test set introduced by Dai \etal \cite{dai2022flare7k}, comprising 100 real-world flare-free images and their corresponding flare-corrupted counterparts.

\vspace{-0.8em}
\subsection{Benchmark Comparison}
We demonstrate the superiority of Difflare by comparing our method both quantitatively and qualitatively with existing benchmark methods. Specifically, we select a night-time dehazing model by Zhang \etal \cite{zhang2018single}, a lens flare removal model with pure synthetic data by Wu \etal \cite{2020How}, a lens flare removal model trained with Flare7K dataset by Zhou \etal \cite{zhou2023improving}, and the state-of-the-art method by Dai \etal \cite{dai2023flare7kpp} using UFormer as backbone. We also present results for simply comparing the input flare-corrupted image its corresponding with flare-free image. For those methods with their pre-trained model released, we test directly with their released code and pre-trained models. For methods lacking publicly released pre-trained models, we follow their training pipeline and adopt our data augmentation settings to retrain their models, ensuring a fair comparison.

\vspace{-0.8em}
\subsubsection{Quantitative Comparison}
For quantitative comparison, we utilize widely accepted full-reference metrics: Peak Signal-to-Noise Ratio (PSNR) and Structural Similarity (SSIM) \cite{article}. Furthermore, given the significance of perceptual quality in image assessment, we employ reference-free metrics MUSIQ \cite{ke2021musiq} and CLIPIQA \cite{wang2022exploring} to gauge the perceptual performance of our results. As shown in the \hyperref[tab:quantitative]{Table 1} below, Since Zhang \etal.’s method \cite{DBLP:journals/corr/abs-2008-03864} is mainly designed
for nighttime haze, it can effectively remove the lens flare, but it also significantly deviates the image's color, leading to a noticable decrease in structural similarity and perceptual quality. Wu \etal's method \cite{2020How} and Zhou \etal's method preserve the light source by adding back the center of the flare with a given threshold, leading to over-sharpen results, thus result in a relatively low perceptual quality. Dai \etal's method \cite{dai2023flare7kpp} achieves the best PSNR, thanks to its data synthesize pipeline and network structure. However, it occasionally removes the light source or produces over-sharpen results, leading to decrease in MUSIQ and CLIPIQA. Our proposed method outperforms the existing benchmark methods in structural similarity preservation, due to the proposed AFFM, and can generate results with best perceptual quality, thanks to the prior captured by PTDM.
\begin{table}[h]
\centering
\label{tab:quantitative}
\begin{tabular}{c|c c c c c c}
\hline
Metrics & Input & Wu  \cite{2020How} & Zhang \cite{DBLP:journals/corr/abs-2008-03864}&Zhou \cite{zhou2023improving}&Dai \cite{dai2022flare7k} & \textbf{Difflare (Ours)} \\
\hline
PSNR $\uparrow$& 22.561 & 24.613 & 21.022 & 25.184 & {\bf26.978} & \underline{26.063}\\
SSIM $\uparrow$& 0.857 & 0.871 & 0.784 & 0.872 & \underline{0.890} & {\bf0.898}\\
MUSIQ $\uparrow$& 59.34 &57.29 & 55.46&\underline{59.09} & 59.03 & \textbf{59.48}\\
CLIPIQA $\uparrow$& 0.332& 0.312 & 0.279 &0.281 & \underline{0.337} & \textbf{0.341}\\
\hline

\end{tabular}
\caption{\small{Quantitative comparison of the SOTA methods, $\uparrow$ indicates that higher is better. {\bf Bold} and \underline{underlined} numbers denotes the first and second best results, respectively.}}
\vspace{-1.5em}
\end{table}

\begin{figure}
\centering
\bmvaHangBox{\includegraphics[width=8cm]{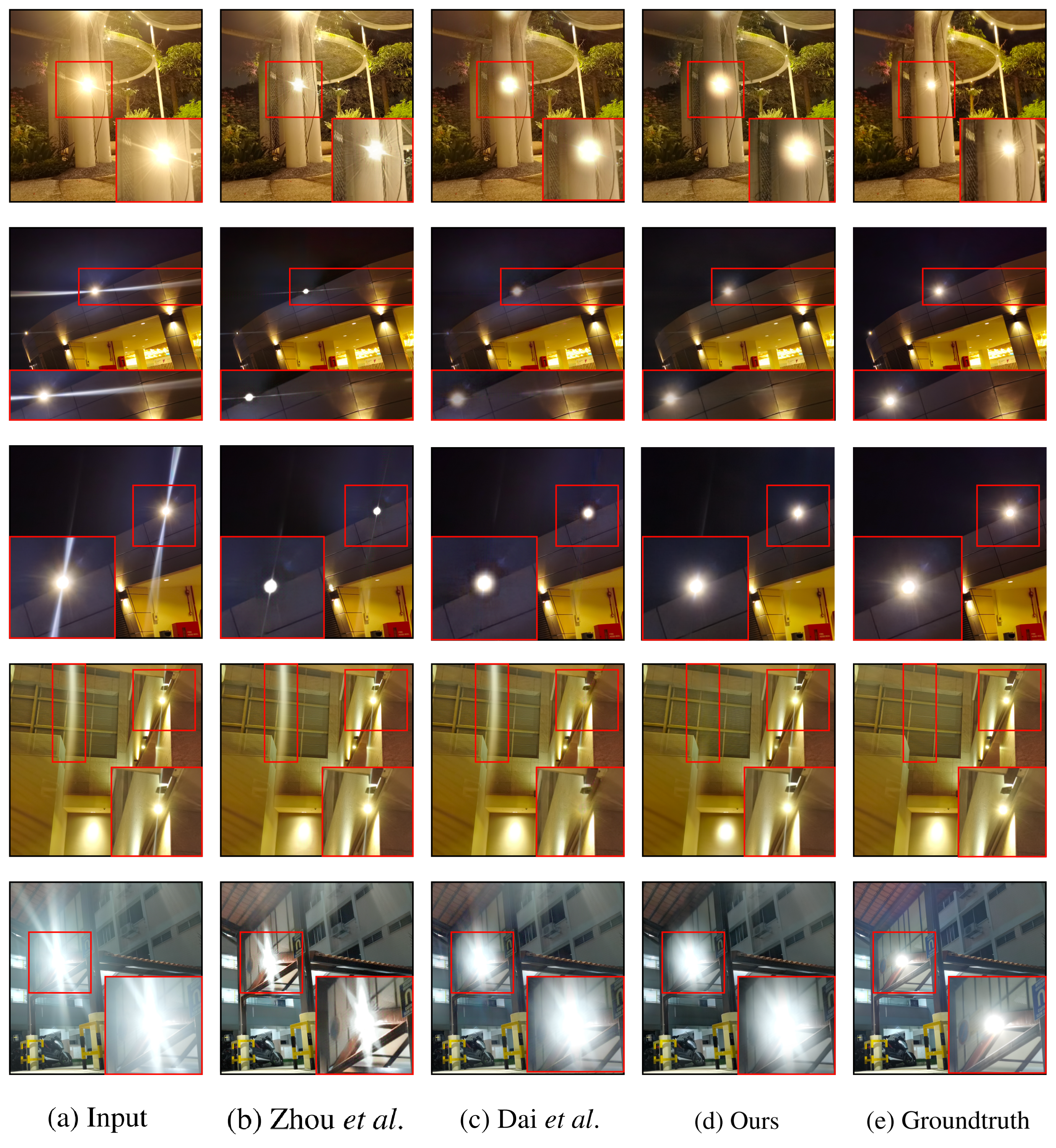}}
\vspace{-0.3em}
\caption{\small{Visual Comparison on Flare7K \cite{dai2022flare7k} testset. Our proposed method can effectively remove lens flare and unwilling artifacts, while harmonizing the recovered light source and the background.}}
\label{figure:qualitative}

\end{figure}

\subsubsection{Qualitative Comparison}

We visually demonstrate the effectiveness of Difflare by juxtaposing its results with those of other benchmark methods in \hyperref[figure:qualitative]{Figure 3}. The figure illustrates that Difflare generates satisfactory results in removing lens flare from flare-corrupted areas. Despite performing lens flare removal in latent space, images generated by Difflare exhibit no difference in flare-free areas, thanks to our proposed AFFM. In rows 1, 2, and 5, Zhou \etal's method  \cite{zhou2023improving} tends to yield overly sharp results, whereas Difflare generates more harmonious results with the background, attributable to PTDM utilization. For rows 1, 2 and 5, it can be seen that the method proposed by Zhou \etal \cite{zhou2023improving} tends to produce over sharp results, while Difflare tends to produce results which are more harmonious with the background, thanks to the use of PTDM. Rows 2 and 4 demonstrate that Difflare better preserves light sources than other methods and effectively eliminates undesirable artifacts caused by lens flare. Row 3 indicates that Difflare produces results more closely resembling the ground truth images.
\begin{table}[t]
\centering
\label{tab:ablation1}
\begin{tabular}{c|c c c c c c}
\hline
Metrics & Ours w/o AFFM & Ours w/ non-guided AFFM  & Ours Full \\
\hline
PSNR $\uparrow$& 18.773 & 25.772 & {\bf26.063}\\
SSIM $\uparrow$& 0.671 & 0.896 & {\bf0.898} \\
MUSIQ $\uparrow$& 56.54 & 58.94 & {\bf59.48}\\
CLIPIQA $\uparrow$& 0.256& 0.307 & {\bf0.341}\\
\hline
\end{tabular}
\vspace{0.8em}
\caption{\small{Quantitative comparison on ablation study of our method. Column 1 represent result of our method with merely SGIM. Column 2 represent result of our method without LGP mask guidance, Column 3 is the result of our full settings.}}

\end{table}

\subsection{Ablation Study}

\begin{figure}
\centering
\bmvaHangBox{\includegraphics[width=10cm]{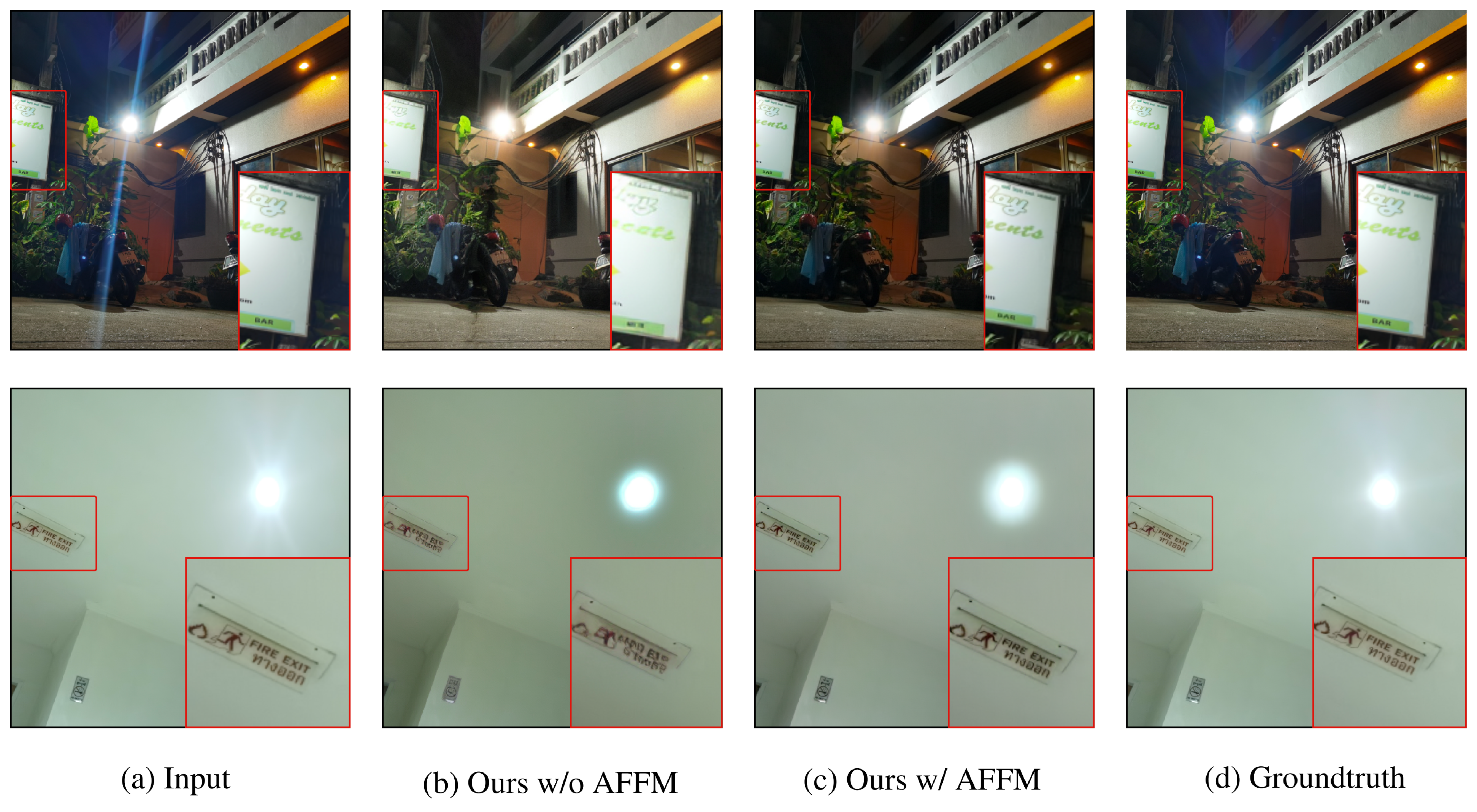}}\\
\caption{\small{Visual comparison on the effect of AFFM. Without AFFM, SGIM can effectively remove lens flare, but there are significant distortions on flare-free areas. When AFFM is employed, the fidelity of flare-free areas has been maintained between the input image and the restored image.}}
\label{fig:ablation1}
\vspace{-1em}
\end{figure}


\vspace{0.2em}
{\bf The effect of AFFM.} We conduct an ablation study to illustrate the importance of our proposed AFFM, comparing our method with and without AFFM. Columns 1 and 3 from \hyperref[tab:ablation1]{Table 2}  demonstrate that the quality of reconstruction results significantly improves with the inclusion of AFFM, as evidenced by the increased PSNR and SSIM values. From \hyperref[fig:ablation1]{Figure 4}, it is obvious that AFFM effectively helps to maintain fidelity in flare-free areas.\vspace{0.4em}\\
{\bf The importance of LGP-based mask in AFFM.}
We explore the effect of the LGP-based mask on the training process of AFFM by training AFFM with and without the guidance of the LGP mask. As shown in \hyperref[tab:ablation1]{Table 2} column 2 and 3, the LGP mask guidance significantly enhances perceptual quality and generates results with improved fidelity. Hence, it demonstrates that our modification to the self-attention mechanism is beneficial for the preservation of fidelity in flare-free areas.
\newpage
\section{Conclusion}
\vspace{-0.6em}
In this paper, we propose a novel approach called Difflare for lens flare removal, which aims to leverage the generative prior captured in pretrained latent diffusion models. Leveraging the properties of diffusion models, we meticulously design a multi-scale guidance injection module and a feature fusion module. Additionally, we consider the optical characteristics of lens flare. Extensive experimentation demonstrates the efficacy of our novel approach in effectively removing lens flare while maintaining high perceptual quality. Our proposed method can be beneficial for more robust applications of high-level computer vision tasks.

\section{Acknowledgement}
This work was supported by National Natural Science Foundation of China under Grant 62306061, and Guangdong Basic and Applied Basic Research Foundation (Grant No. 2023A\\1515140037)

\bibliography{egbib}
\end{document}